\documentclass[10pt,twocolumn,letterpaper]{article}

\usepackage{cvpr}              

\usepackage{graphicx}
\usepackage{amsmath}
\usepackage{amssymb}
\usepackage{booktabs}
\usepackage[frozencache,cachedir=.]{minted}
\usepackage{listings}
\usepackage{array}

\usepackage[pagebackref,breaklinks,colorlinks]{hyperref}

\usepackage[capitalize]{cleveref}
\crefname{section}{Sec.}{Secs.}
\Crefname{section}{Section}{Sections}
\Crefname{table}{Table}{Tables}
\crefname{table}{Tab.}{Tabs.}

\def\cvprPaperID{10} 
\def\confName{CVPR}
\def\confYear{2022}

\begin{document}
\renewcommand{\lstlistingname}{Algorithm}

\title{Feature Refinement to Improve High Resolution Image Inpainting}

\author{Prakhar Kulshreshtha \thanks{authors contributed equally}\\
Geomagical Labs, Inc.\\
{\tt\small prakhar@geomagical.com}
\and
Brian Pugh$^*$\\
Geomagical Labs, Inc.\\
{\tt\small bpugh@geomagical.com}
\and
Salma Jiddi\\
Geomagical Labs, Inc.\\
{\tt\small salma@geomagical.com}
}



\maketitle


\begin{abstract}
In this paper, we address the problem of degradation in inpainting quality of neural networks operating at high resolutions.
Inpainting networks are often unable to generate globally coherent structures at resolutions higher than their training set. 
This is partially attributed to the receptive field remaining static, despite an increase in image resolution.
Although downscaling the image prior to inpainting produces coherent structure, it inherently lacks detail present at higher resolutions.
To get the best of both worlds, we optimize the intermediate featuremaps of a network by minimizing a multiscale consistency loss at inference.
This runtime optimization improves the inpainting results and establishes a new state-of-the-art for high resolution inpainting.
Code is available at: \footnotesize{\url{https://github.com/geomagical/lama-with-refiner/tree/refinement}}.
\end{abstract}

\section{Introduction}
\label{sec:intro}

Image inpainting is the task of filling missing pixels or regions in an image \cite{deepfillv1}.
This task finds application in image restoration, image editing, Augmented Reality, and Diminished Reality\cite{shoei_2017}\cite{bardi_2016}.
Several methods have been proposed to solve this problem.
\cite{bartelmio, telea} inpaint missing regions using gradient guided diffusion of colors from neighboring pixels.
\cite{exemplarbased, patchmatch} sample patches from unmasked areas of the image that satisfy well-defined similarity criteria.
Patch-based solutions are widely adopted in image editing tools like Gimp\cite{gimp} and Photoshop\cite{photoshop}.

Existing approaches often struggle with global consistency when the masked-region is large enough to encompass multiple texture or semantic regions \cite{deepfillv2, comodgan}. 
Conditional Generative Adversarial Networks (cGAN) have been developed to address this issue via an intermediate global representation \cite{contextencoder,partialconv, madf,deepfillv1, deepfillv2}.
Even with cGAN, a large receptive field is critical for high inpainting performance \cite{lama}.
Various techniques have been proposed to increase the effective receptive field, such as Fourier convolutions\cite{lama}, diffusion models \cite{latentdiffusion}, contextual transformations \cite{aotgan}, and transformers \cite{zits, mat}.

\begin{figure}[htb!!!]
    \centering
    \includegraphics[width=0.9\columnwidth]{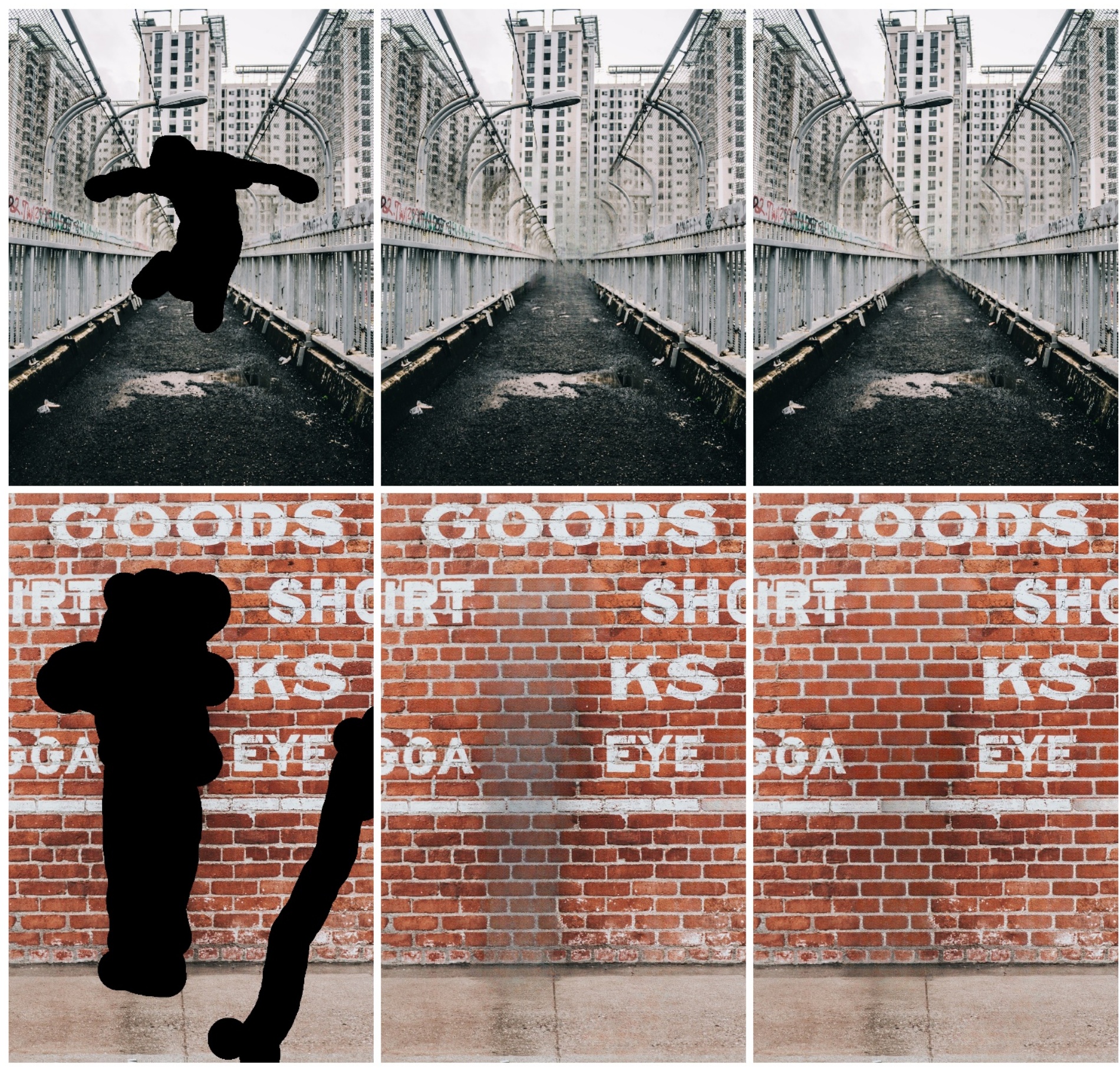}
    \caption{Results from our multiscale refinement. Left: input image. Center: inpainting with Big-LaMa\cite{lama}. Right: inpainting with Big-LaMa + our refinement.}
    \label{fig:qualitative}
\end{figure}

In this work, we focus on improving the inpainting quality of existing networks at high resolutions.
Increasing the operating image-size proportionally decreases the available local context to the network when inpainting a region, which causes incoherent structures and blurry textures \cite{aotgan}. 
To solve this problem, we propose a novel coarse-to-fine iterative refinement approach that optimizes featuremaps via a multiscale loss. 
By using lower resolution predictions as guidance, the refinement process produces detailed high resolution inpainting results while maintaining the color and structure from low resolution predictions (Fig. \ref{fig:qualitative}).
No additional training of the inpainting network is required; only featuremaps are refined during inference \cite{adnn, pnp}.
\vspace{-0.25cm}

\section{Multiscale Feature Refinement}

\begin{figure}[htb!]
    \centering
    \includegraphics[width=\columnwidth]{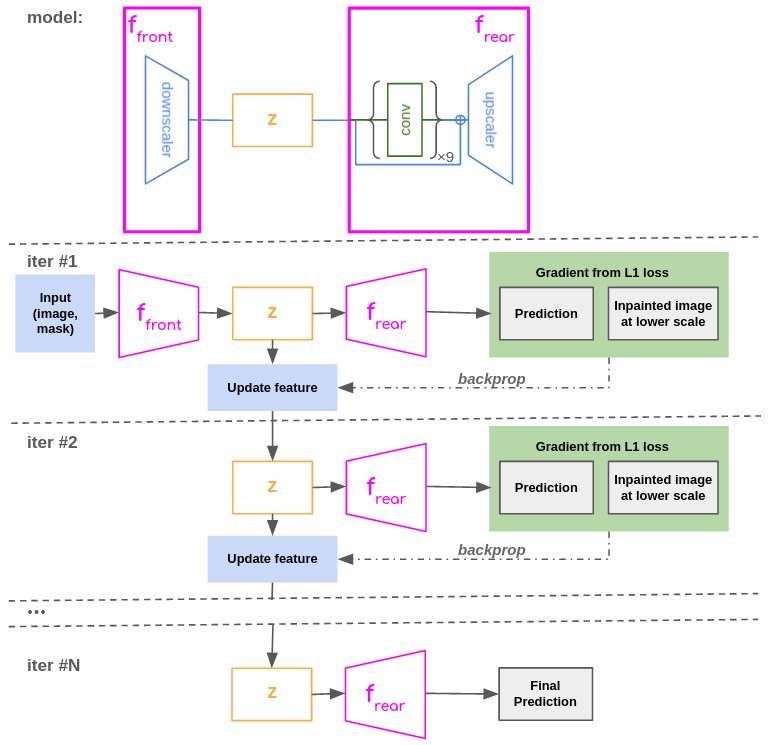}
    \caption{Iterative refinement of the inpaint prediction at a single scale by optimizing the latent featuremap $z$.}
    \label{fig:refinement}
\end{figure}

Our multiscale feature refinement follows a coarse-to-fine approach to iteratively add more detail to an inpainting prediction (Fig. \ref{fig:refinement}).

An image-pyramid of the input RGB image and inpainting mask is constructed to be used as network inputs at multiple inference resolutions.
The smallest scale is approximately equal to the network's training resolution.
We assume that the network will perform best at training resolution and use it as the basis for all inpainting structure guidance. 

The model is split into ``front" and ``rear" sections, similar to \cite{pnp}.
Typically, these correspond to the encoder and decoder portions of the network, respectively.
At the lowest resolution, we perform a single forward pass through the entire inpainting model to get an initial inpainting prediction.
For each subsequent scale, we run a single forward pass through the ``front" to generate an initial featuremap $z$.
Multiple featuremaps (e.g. from skip-connections) can be jointly optimized, but are not investigated in this paper.

The rear part of the network processes $z$ to produce an inpaint prediction.
The prediction is then downscaled to match the resolution of the previous scale's result.
Downscaling involves applying a Gaussian filter, followed by bilinear interpolation.
The Gaussian filter removes high frequency components and prevents aliasing during downscaling. 
An L1 loss is computed between the masked inpainted regions and is minimized by updating $z$ via backpropagation.
This will optimize $z$ to produce a higher resolution prediction that has similar characteristics to the previous scale.

\begin{figure}[ht!]
    \centering
    \includegraphics[width=6.5cm]{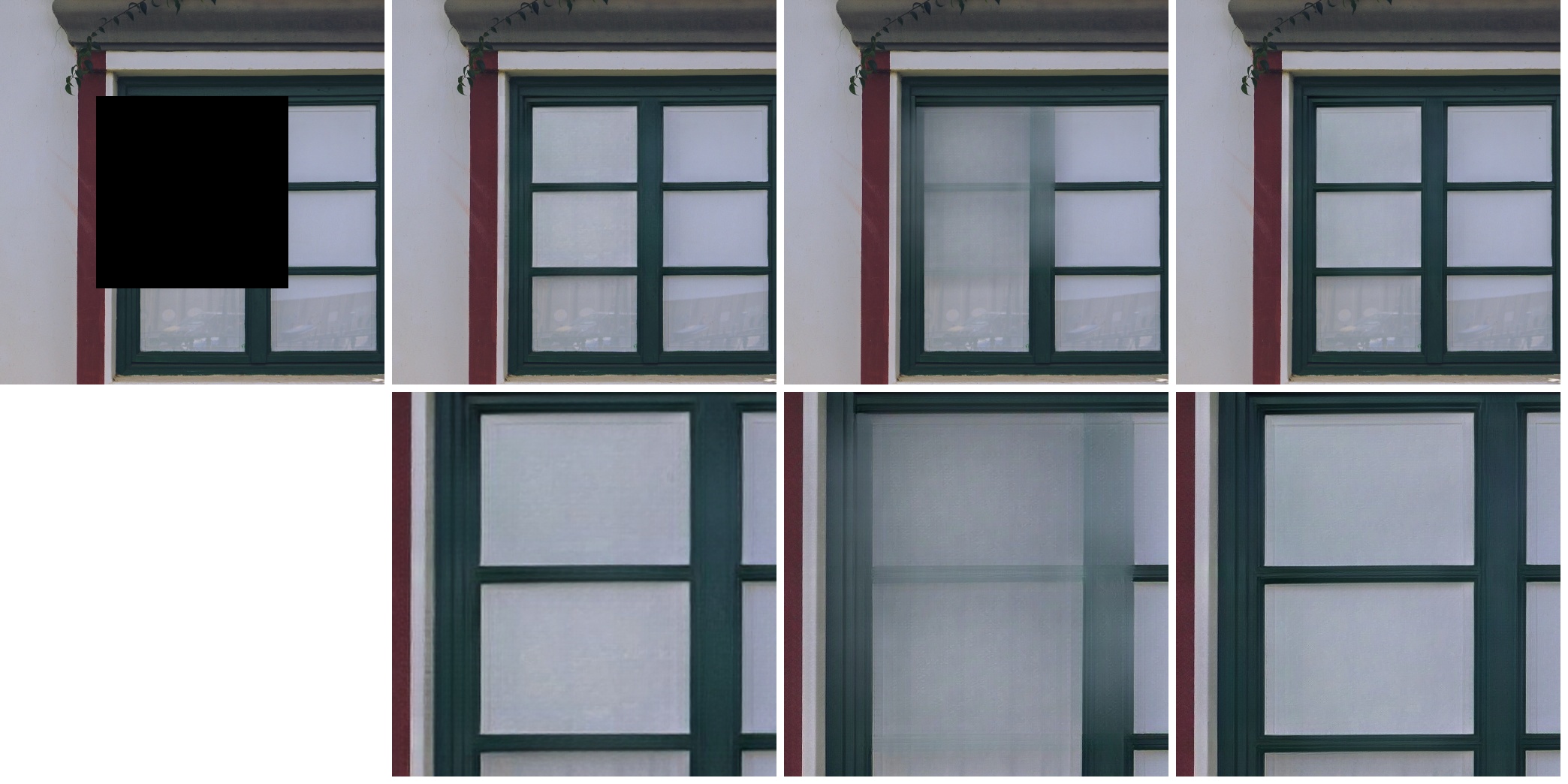}
    \scriptsize
    \begin{tabular}{>{\centering\arraybackslash}p{1.2cm}>{\centering\arraybackslash}p{1.2cm}>{\centering\arraybackslash}p{1.2cm}>{\centering\arraybackslash}p{1.2cm}}
        Masked input & Infill@512 & Infill@1024 & Infill@1024 (refined)\\
    \end{tabular}
    \caption{From left to right, first row: (i) input image, (ii) inpainting at size 512, (iii) inpainting at size 1024 (iv) inpainting at 1024 with refinement.  Second row: zoomed-in corresponding inpainted areas.}
    \label{fig:house}
\end{figure}

Figure \ref{fig:house} shows an example of how refinement improves the quality of predictions.
At high resolutions (1024px), our method has significant improvements regarding structure-completion when compared to Big-LaMa \cite{lama}.
Our refinement also contains more details compared to the upscaled low resolution prediction (512px).

Python pseudocode for our multiscale refinement is described in Algorithm \ref{multiscale-refinement}. The \verb{multiscale_inpaint{ function generates the image pyramid and iterates over the multiple scales, while \verb{predict_and_refine{ produces a per-scale refined prediction.

\begin{table*}[htb!!!]
    \centering
    \small
    \begin{tabular}{c|cc|cc|cc|c}
        Method & \multicolumn{2}{c|}{Thin Brush} & \multicolumn{2}{c|}{Medium Brush} & \multicolumn{2}{c|}{Thick Brush} & {Time per Image}\\
        \cline{2-8}
         & FID$\downarrow$ & LPIPS$\downarrow$ & FID$\downarrow$ & LPIPS$\downarrow$ & FID$\downarrow$ & LPIPS$\downarrow$ & Seconds \\
        \hline
        AOTGAN\cite{aotgan} & 17.387 & 0.133 & 34.667 & 0.144 & 54.015 & 0.184 & 0.43 \\
        LatentDiffusion\cite{latentdiffusion} & 18.505 & 0.141 & 31.445 & 0.149 & 38.743 & 0.172 & 31.56\\
        MAT\cite{mat} & 16.284 & 0.137 & 27.829 & 0.135 & 38.120 & 0.157 & 0.56\\
        ZITS\cite{zits} & 15.696 & 0.125 & 23.500 & 0.121 & 31.777 & 0.140 & 4.14\\
        LaMa-Fourier\cite{lama} & 14.780 & 0.124 & 22.584 & 0.120 & 29.351 & 0.140 & \bf{0.16}\\
        Big-LaMa\cite{lama} &  \textbf{13.143} & 0.114 & 21.169 & 0.116 & 29.022 & 0.140 & 0.26\\
        \bf{Big-LaMa+refinement (ours)} & 13.193 & \textbf{0.112} & \textbf{19.864} & \textbf{0.115} & \textbf{26.401} & \textbf{0.135} & 4.56\\
    \end{tabular}
    \caption{Performance comparison against recent inpainting approaches on 1k 1024x1024 size images sampled from \cite{unsplash}. Inference time per image was calculated on a single NVIDIA RTX A5000 GPU.}
    \label{tab:my_label}
\end{table*}

\begin{listing}[!htb]
\begin{minted}[fontsize=\scriptsize]{python}
def predict_and_refine(image, mask, inpainted_low_res, 
      model, lr=0.001, n_iters=15):
  z = model.front.forward(image, mask)
  # configure optimizer to update the featuremap
  optimizer = Adam([z], lr) 
  for _ in range(n_iters):
    optimizer.zero_grad()
    inpainted = model.rear.forward(z)
    inpainted_downscaled = downscale(inpainted)
    loss = l1_over_masked_region(
      inpainted_downscaled, inpainted_low_res, mask
    )
    loss.backward()
    optimizer.step()  # Updates z
  # final forward pass
  inpainted = f_rear.forward(z)
  return inpainted

def multiscale_inpaint(image, mask, model, smallest_scale=512):
  images, masks = build_pyramid(image, mask, smallest_scale)
  n_scales = len(images)
  # initialize with the lowest scale inpainting
  inpainted = model.forward(images[0], masks[0])
  for i in range(1, n_scales):
    image, mask = images[i], masks[i]
    inpainted_low_res = inpainted
    inpainted = predict_and_refine(
      image, mask, inpainted_low_res, model
    )
  return inpainted
\end{minted}
\caption{PyTorch pseudocode of multiscale refinement.}
\label{multiscale-refinement}
\end{listing}
\pagebreak{}

\begin{figure}[b]
    \centering
    \includegraphics[width=7.5cm]{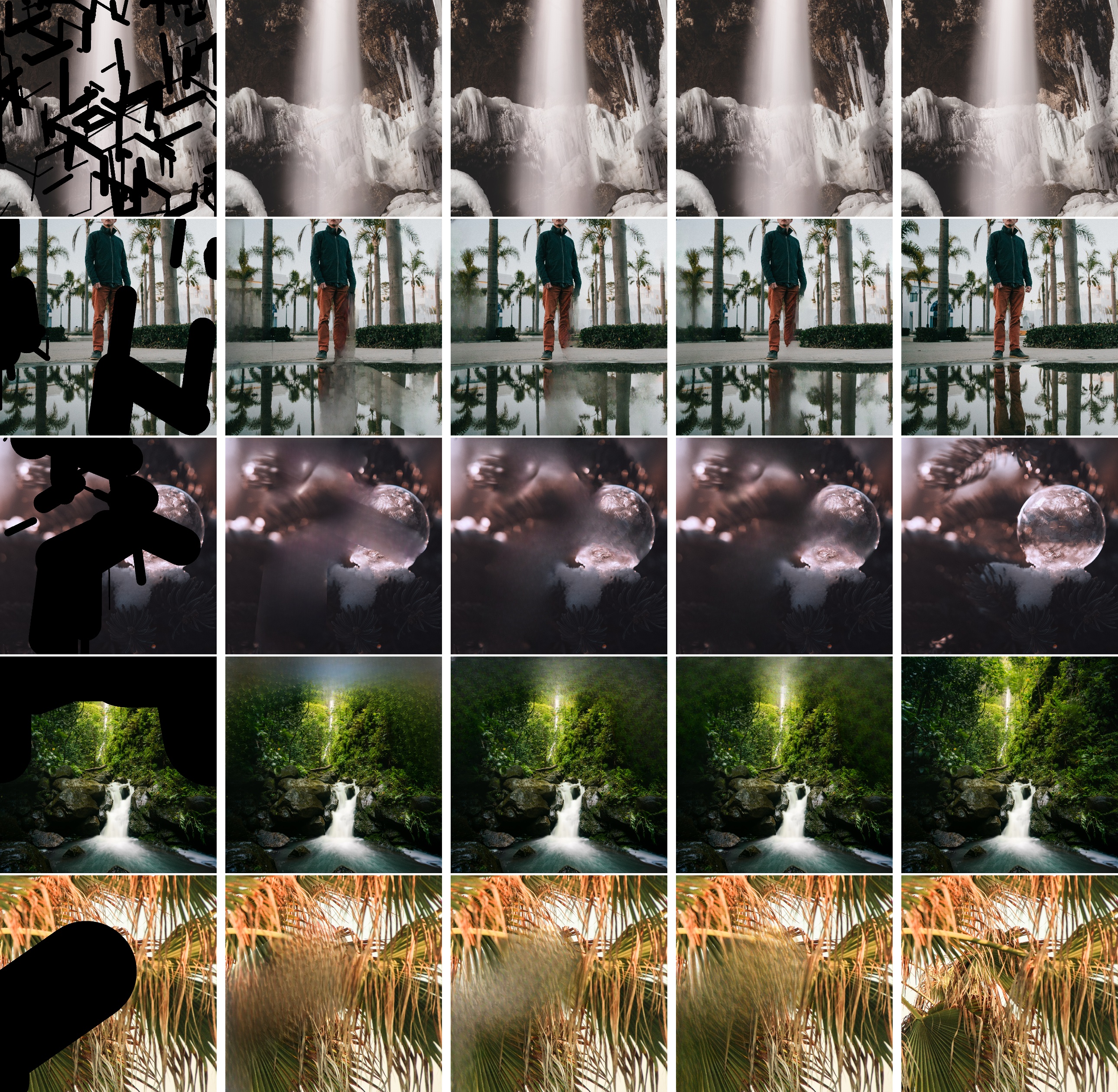}
    \scriptsize
    \begin{tabular}{>{\centering\arraybackslash}p{1.06cm}>{\centering\arraybackslash}p{1.06cm}>{\centering\arraybackslash}p{1.06cm}>{\centering\arraybackslash}p{1.06cm}>{\centering\arraybackslash}p{1.06cm}}
        Input & ZITS\cite{zits} & LaMa\cite{lama} & LaMa+refine \bf{(ours)} & Ground truth\\
    \end{tabular}

    \caption{Comparison of our refinement results against several recent state-of-the-art methods. Best viewed digitally.}
    \label{fig:testset}
\end{figure}

\section{Experiments}

In our experiments, we apply iterative multiscale refinement to Big-LaMa\cite{lama}.
A downscaling factor of 2 is used to build the image pyramid. 
The output featuremap from the downscaler portion of Big-LaMa (see $f_{front}$ in Fig. \ref{fig:refinement}) will be optimized.
This featuremap was chosen based on the observation that featuremaps farther from the prediction layer have a larger receptive field and are able to influence more of the output\cite{pnp}.

At each scale, we perform $15$ refinement iterations using Adam optimizer with a learning rate of $0.002$.
To prevent the network from optimizing against low resolution infill in thin regions where the network is already performing well, we erode the mask with a $15$ pixel circular kernel prior to applying L1 loss to the inpainted regions.

\section{Results}

Inpainting networks are typically benchmarked on Places2 \cite{places}.
However, this dataset does not have high resolution images for evaluation purposes.
Instead, we will use images from the Unsplash-Lite Dataset, which contains 25k high resolution nature-themed photos \cite{unsplash}.
We randomly sampled 1000 images to evaluate on (linked \href{https://tinyurl.com/mpr7w3xt}{\textcolor{blue}{[here]}}).

Each image is resized and cropped to $1024$x$1024$, and a set of masks is generated with thin, medium, and thick brush strokes, using the methodology described in \cite{lama}.
These different mask-types are evaluated separately to observe the effect the width of the mask has on image inpaint quality. 
In accordance with recent works \cite{lama, zits}, performance is evaluated using FID scores \cite{fid} and LPIPS \cite{lpips}. 

\begin{figure*}
    \centering
    \includegraphics[width=\textwidth]{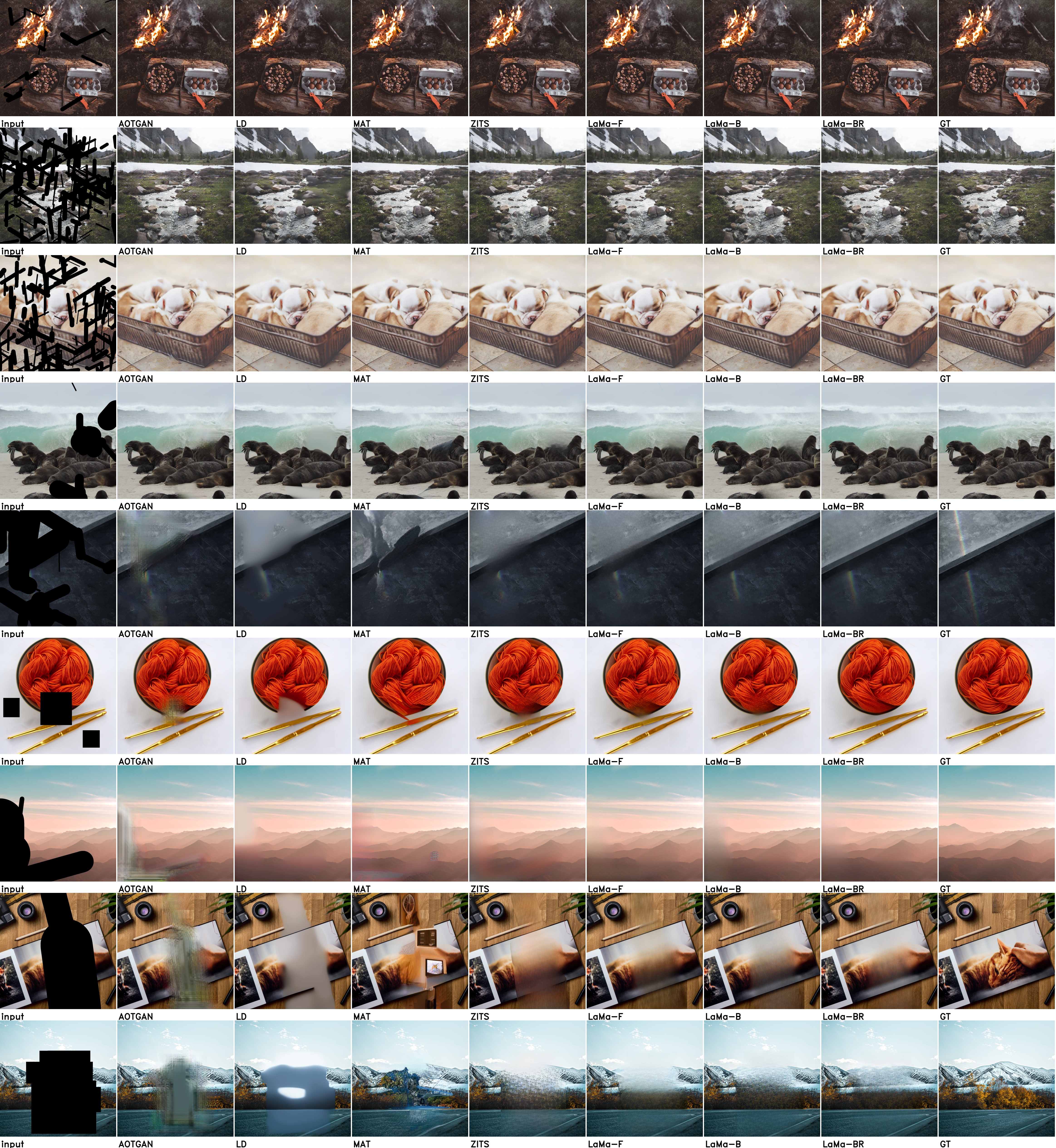}
    \caption{Comparison on more examples of size 1024x1024 from Unsplash dataset\cite{unsplash}. In each row, from left to right, we show: input image, output using AOTGAN\cite{aotgan}, Latent-Diffusion(LD)\cite{latentdiffusion}, MAT\cite{mat}, ZITS\cite{zits}, LaMa-Fourier(LaMa-F)\cite{lama}, Big-LaMa(LaMa-B)\cite{lama}, Big-LaMa with refinement(LaMa-BR - ours). Last image in each row is the ground-truth (GT).}
    \label{fig:cvpr_methods}
\end{figure*}

We compare against other methods in Table \ref{tab:my_label} and Figures \ref{fig:testset}, \ref{fig:cvpr_methods}.
Our method outperforms reported state-of-the-art inpainting networks for medium and thick masks, while performing similarly to Big-LaMa \cite{lama} for thin masks. 
The thin-mask performance is similar because there is sufficient surrounding context to complete the structure.

Although our refinement produces higher scoring results, it also takes significantly longer to process an image.
For each image, multiple forward and backward passes are required. This increase inference-time proportionally with number of scales and optimization steps.
Refinement also increases memory usage because gradients are required at runtime, consequently reducing the maximum resolution that can fit in GPU memory. Our approach produces infills with stronger global consistency and sharper textures. Additional results are available via this linked \href{https://youtu.be/gEukhOheWgE}{\textcolor{blue}{video}}.

\section{Conclusion}
We proposed a multiscale refinement technique to improve the inpainting performance of neural networks on images at resolution higher than the native training resolution.
This refinement is network agnostic, and requires no additional model retraining.
Our results indicate that this technique significantly outperforms other state-of-the-art approaches at high resolution inpainting.


\pagebreak
{\small
\bibliographystyle{ieee_fullname}
\bibliography{egbib}
}

\end{document}